# Survey On The Estimation Of Mutual Information Methods as a Measure of Dependency Versus Correlation Analysis

D. Gencaga[a,*], N. Malakar[a], D. J. Lary[a]

[a]*William B. Hanson Center for Space Sciences, University of Texas at Dallas, 800W. Campbell Road, MS/WT15, Texas, 75080 USA*

**Abstract.** In this survey, we present and compare different approaches to estimate Mutual Information (MI) from data to analyse general dependencies between variables of interest in a system. We demonstrate the performance difference of MI versus correlation analysis, which is only optimal in case of linear dependencies. First, we use a piece-wise constant Bayesian methodology using a general Dirichlet prior. In this estimation method, we use a two-stage approach where we approximate the probability distribution first and then calculate the marginal and joint entropies. Here, we demonstrate the performance of this Bayesian approach versus the others for computing the dependency between different variables. We also compare these with linear correlation analysis. Finally, we apply MI and correlation analysis to the identification of the bias in the determination of the aerosol optical depth (AOD) by the satellite based Moderate Resolution Imaging Spectroradiometer (MODIS) and the ground based AErosol RObotic NETwork (AERONET). Here, we observe that the AOD measurements by these two instruments might be different for the same location. The reason of this bias is explored by quantifying the dependencies between the bias and 15 other variables including cloud cover, surface reflectivity and others.

**Keywords:** Information theory, Entropy, Mutual Information, Correlation, Measures of Statistical Dependency, Bayesian data analysis, Uncertainty.
**PACS:** 02, 05, 89.

## INTRODUCTION

We propose a general Bayesian approach to estimate Mutual Information (MI) from data to analyse statistical dependencies between different variables of interest in a system. It is known that statistical independence implies uncorrelatedness, but the reverse is only guaranteed for jointly Gaussian distributed variables. As the correlation coefficient is estimated using the first and second order statistics in a linear way, higher order dependencies cannot be investigated by this approach. MI, on the other hand, is defined through the probability distributions, utilizing the whole statistical dependencies [1]. Even though this measure can be used to estimate the aforementioned dependencies perfectly in theory [2], different approaches have been developed in the literature to estimate it from data as accurate as possible, because bias and variance differences arise due to each estimation technique.

Fixed bin-width histogram based methods are widely used in the literature due to their computational efficiency [3]. However, as the number of unfilled bins increase with the increasing number of data dimensions, variable bin-width histograms have been proposed as alternatives [4-6]. As the bin heights are directly related to the highly occupied data sections, there is no drawback of having unfilled bins in these methods. Although these methods have been used satisfactorily, they all have different bias and variances in the eventual information-theoretic quantity estimate. Moreover, as these methods utilize point estimations, we cannot report uncertainties around the estimates.

Here, we extend the piece-wise constant Bayesian methodology in [7] using a general Dirichlet prior in a similar way that is utilized to estimate different information-theoretic quantities like the transfer entropy in [8]. In this estimation method, we use a two-stage approach where we approximate the probability distribution first and then calculate the marginal and joint entropies to be substituted in the MI formula estimation [7,8]. Here, we demonstrate the performance of this Bayesian approach versus the others for computing the dependency between different variables. We also compare these with linear correlation analysis.

Finally, we apply MI and correlation analysis to the identification of the bias in the determination of the aerosol optical depth (AOD) by the satellite based Moderate Resolution Imaging Spectroradiometer (MODIS) and the ground based AErosol RObotic NETwork (AERONET). Here, we observe that the AOD measurements by these two instruments might be different for the same location. The reason of this bias is explored by quantifying the dependencies between the bias and 15 other variables including cloud cover, surface reflectivity and others.

The rest of the text continues with a brief background on information-theoretic quantities and different methods of estimating them from data. After presenting the technical details of our method in Section 3, we demonstrate the comparison results in Section 4 on a linearly coupled autoregressive model. Conclusions are drawn in Section 5.

## ESTIMATION OF INFORMATION-THEORETIC QUANTITIES FROM DATA

In the literature, the correlation coefficient has been used to determine the statistical dependencies because of its computational efficiency and simplicity. The correlation coefficient is a linear measure and it is defined by the following formula:

$$\rho(X,Y) = \frac{E[(X-\mu_X)(Y-\mu_Y)]}{\sigma_X \sigma_Y} \quad (1)$$

where the numerator denotes the covariance between X and Y. $\sigma_X$ and $\sigma_Y$ represent the standard deviations of X and Y, respectively; whereas $\mu_X$ and $\mu_Y$ denote the mean values. E[.] is the expectation operator and defined as follows:

$$E[X] = \sum_{x \in X} x p(x) \quad (2)$$

where p(x) is the probability distribution of X. Thus, E[.] is a *linear* averaging operator and the correlation coefficient only utilizes up to the second order statistics.

MI, on the other hand, is defined through the Kullback Leibler divergence of the joint distribution to the product of its marginals as illustrated below:

$$I(X,Y) = \sum_{x \in X} \sum_{y \in Y} p(x,y) \log \frac{p(x,y)}{p(x)p(y)} \quad (3)$$

which is zero when X and Y are statistically independent, i.e. $p(x,y) = p(x)p(y)$. As (3) is not bounded from above, normalized versions have been proposed in the literature [8,9]. For continuous variables, a normalized MI is given by

$$L(X,Y) = \left[1 - \exp\{-2I(X,Y)\}\right]^{1/2} \quad (4)$$

which normalizes the MI between 0 and 1, meaning that there is no dependency and that Y is described totally having observed X, respectively. (3) can be rewritten in terms of Shannon entropies as given below:

$$I(X,Y) = H(X) + H(Y) - H(X,Y) \quad (5)$$

where H(X) and H(X,Y) denote the marginal and joint entropies, respectively. These are estimated from data using the following definitions [2]:

$$H(X) = -\sum_{x \in X} p(x) \log p(x) \qquad H(X,Y) = -\sum_{x \in X} \sum_{y \in Y} p(x,y) \log p(x,y) \quad (6)$$

denoting the marginal and the joint entropies.

**Generalized Bayesian Piece-wise Constant Model for Entropy Estimation**

Here, we notice that the Shannon entropies need to be estimated prior to the estimation of the MI. To estimate these entropies, we need to compute the probability distributions, p(.), as illustrated in (6). In the literature, these distributions are generally approximated by histograms [3]. However, detecting the optimal number of bins has always been a challenge for an accurate estimate [10]. In [10], the probability distribution is defined as a piece-wise constant model and each bin is attained a priori probability distribution, making it a Bayesian model, rather than a frequentist histogram. Each piece-wise constant portion, approximating a continuous probability density function (pdf), has the following Dirichlet pdf [10], where $\beta$ is set to 0.5.

$$p(\pi \mid M) = \frac{\Gamma\left(\sum_{i=1}^{M} \beta\right)}{\prod_{i=1}^{M} \Gamma(\beta)} \left[\pi_1 \pi_2 ... \pi_{M-1} \left(1 - \sum_{i=1}^{M-1} \pi_i\right)\right]^{\beta-1} \quad (7)$$

where the Dirichlet exponent is $\beta$. The following uniform prior is used for the number of bins:

$$p(M) = \begin{cases} C^{-1}, & \text{if } 1 \leq M \leq C \\ 0, & \text{otherwise} \end{cases} \quad (8)$$

To emphasize the difference between the piece-wise constant, Bayesian model and the widely known frequentist histogram; the following figure is shown. Below, the bin heights, having their own pdf's, are denoted by $\pi_k$ for the k$^{th}$ bin, whereas the heights of each bin in a regular histogram are just the frequencies of data.

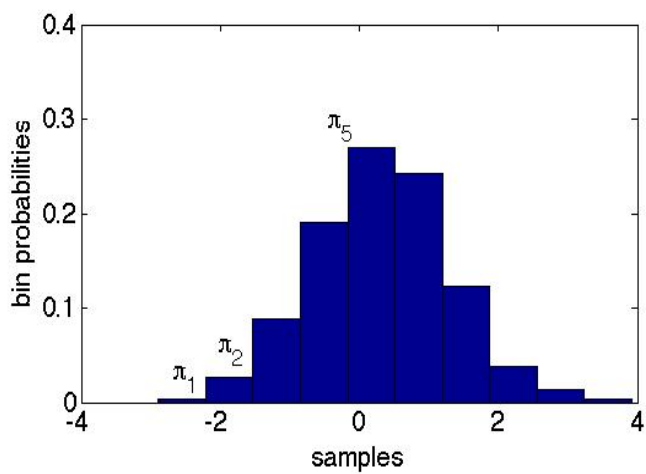

FIGURE 1. Piece-wise constant pdf model with bin heights are illustrated with pi's and range is divided into M=10 bins

In this method, each of $N$ observed datum point is placed into one of $M$ fixed width bins. If the volume and the bin probabilities of each multivariate bins are denoted by V and $\pi_k$ for the k$^{th}$ bin, respectively, then the likelihood of the data can be given by the following multinomial distribution [11]:

$$p(\mathbf{d} \mid M, \pi) = \left(\frac{M}{V}\right)^N \pi_1^{n_1} \pi_2^{n_2} ... \pi_M^{n_M} \quad (9)$$

where

$$\mathbf{d} = \{d_1,...,d_N\} \quad \text{and} \quad \pi = \{\pi_1,...,\pi_{M-1}\}, \quad \pi_M = 1 - \sum_{k=1}^{M-1} \pi_k$$

In this method, the key point is that the $\beta = 0.5$ in [10] is *generalized to be any value larger than zero*. In [8], it is shown that other $\beta \neq 0.5$ values can provide better entropy and thus MI estimates. We will call this to be the generalized optBINS method as we can also find the optimal number of bins by a similar algebra and estimate it as follows [8]:

$$\hat{M} = \arg\max_{M}\{\log p(M \mid \mathbf{d})\} \qquad (10)$$

where the log posterior pdf of the number of bins is given as follows:

$$\log p(M \mid \mathbf{d}) = N\log M + \log\Gamma(M\beta) - M\log\Gamma(\beta) - \log\Gamma(N + M\beta) + \sum_{k=1}^{M}\log\Gamma(n_k + \beta) + K$$

(11)

In summary, first we form the Dirichlet prior for the bin heights (which are modeled probabilistically, thus we can sample from them) and the uniform prior for the number of bins. As the Dirichlet prior is conjugate of the multinomial likelihood function in (9); after the Bayes formula, the posterior distribution is obtained to be a Dirichlet as well [7]. Then, we can sample from this posterior and obtain numerous values to estimate the mean and the error bars of our entropy estimates.

**Variable bin-width histogram method**

It is known that the fixed bin-width histograms are not so efficient to estimate the pdf's as the dimension of the data increases [7]. Lots of bins tend to become empty, causing bias problems. On the other hand, if the bins are arranged in such a way that the probability masses in each bin are taken to be the same, we can overcome the unfilled bin problem [4,5]. Thus, the observation space is adaptively partitioned into non-overlapping cuboids, having varying sizes, in the multidimensional space. The reader is referred to [5] and the references therein for a better visualization of this method. In this approach, we utilize a chi-square test to form conditionally independent cells [4,5]. At the end, finer resolution is used to describe higher MI regions, whereas lower one is used for less.

**EXPERIMENTS**

As a demonstration and a comparison with the performance of the methods, we work on linearly coupled autoregressive bivariate signals shown below:

$$\begin{aligned} y(i+1) &= 0.5 y(i) + n_1(i) & n_1 &\sim N(0,1) \\ x(i+1) &= 0.6 x(i) + e y(i) + n_2(i) & n_2 &\sim N(0,1) \\ & & e &\in [0.01, 1] \end{aligned} \qquad (12)$$

where $N$ and $e$ denote the normal distribution and the coupling coefficient, respectively. Here, we take three approaches to estimate the MI between x and y: 1. Histogram method with arbitrarily selected number of bins; 2. Generalized optbins with $\beta = 0.05$ and $\beta = 0.5$; 3. Variable bin-width histograms. In Fig. 2., MI estimations are illustrated for different methods. Here, we observe that the frequentist histogram approach with M=30 bins in each direction has the highest bias. On the other hand, Generalized Bayesian piece-wise constant model seems to provide lower bias when the Dirichlet coefficient is smaller. It is also observed that the variable bin-width method behaves similarly to the general optBINS method with lower exponent. The role of the Dirichlet exponent, $\beta$, on the estimation performance is demonstrated in detail in [8].

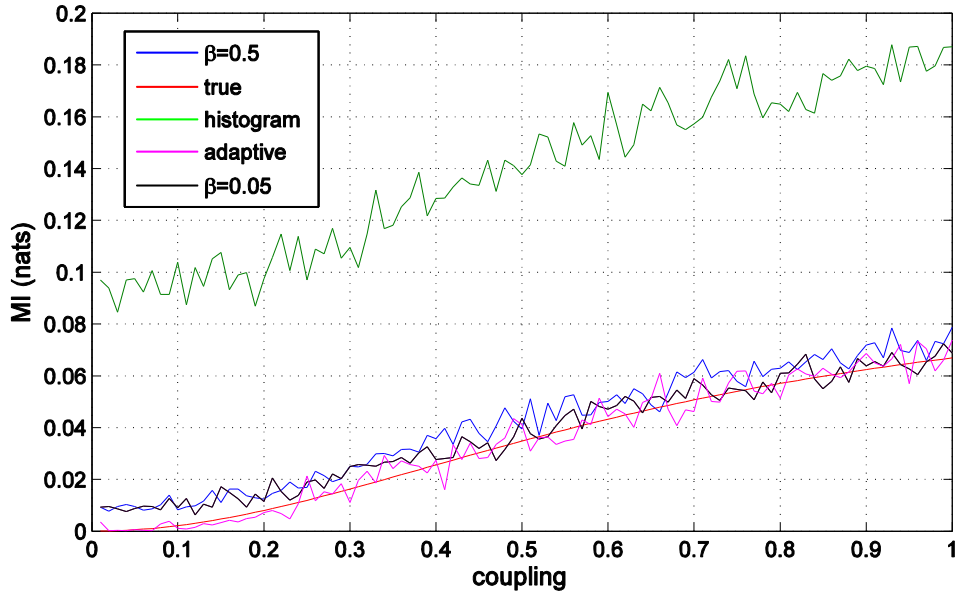

FIGURE 2. Comparison of different MI estimation for the coupled AR signals in (12) as a function of the coupling coefficients

We have just illustrated that the Bayesian method with $\beta = 0.05$ and the adaptive method behave similarly, and their performance is close to the true, analytical curve. Thus, we applied these two methods to explain the differences between the AOD measurements of the MODIS instrument on the NASA Terra and Aqua satellites and the ground based AERONET. Below, we tabulated the MI versus correlation coefficient between the bias and 15 other variables measured by the AERONET. We define the bias for the same location as the difference between the AOD's at 550nm as follows:

Bias = MODIS AOD at 550nm – AERONET AOD at 550nm

We explore the statistical dependencies between this bias term and the following variables: These variables are AERONET's AOD at 550nm, AOD at 470nm, AOD at 660nm, Mean Reflectance (mref) at 470nm, mref at 550nm, Surface Reflectance at 660nm, (surfre0660), 470nm (surfre0470), 2100nm (surfre2100), cloud fraction from land aerosol cloud mask (cfrac), Quality Assurance (QAavg), Solar Zenith, Solar Azimuth, Sensor Zenith, Sensor Azimuth and the ScatteringAngles. Our estimation resulted in the following figures, where the dependencies are measured by the aforementioned MI methods and the correlation coefficient. The components are ranked according to a descending effect in the bias as illustrated in the following figure for 3 different dependency estimation methods:

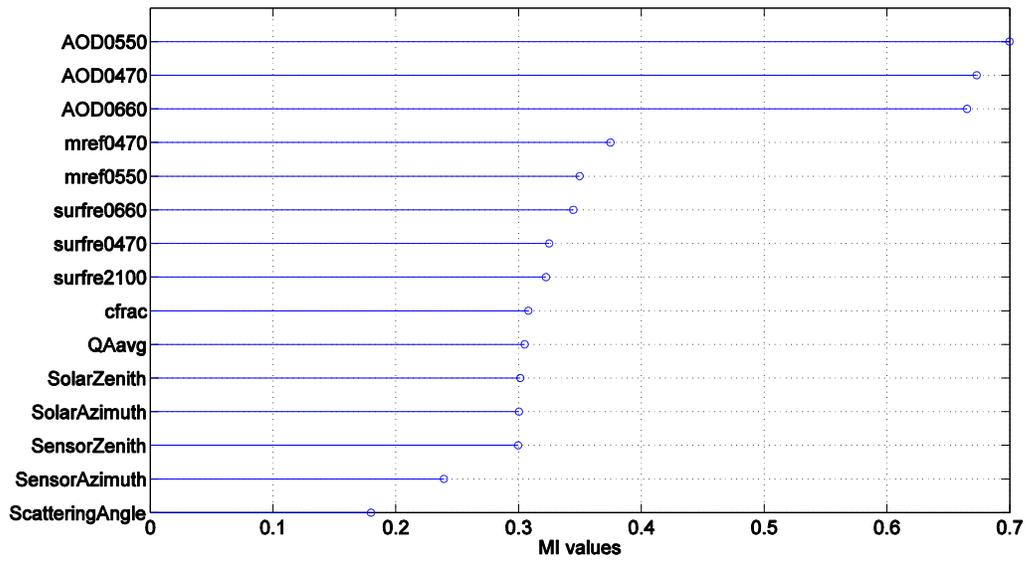

a)

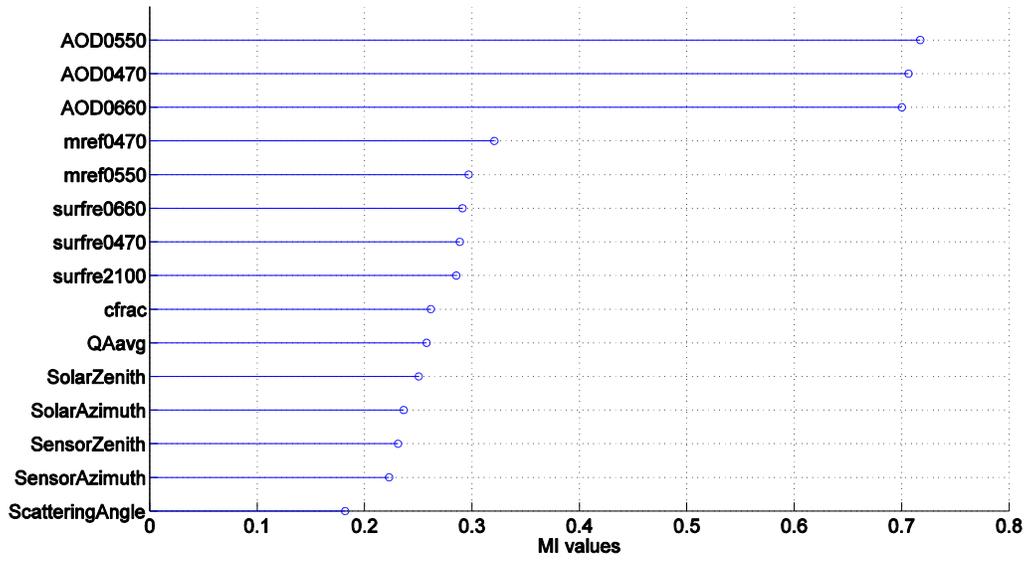

b)

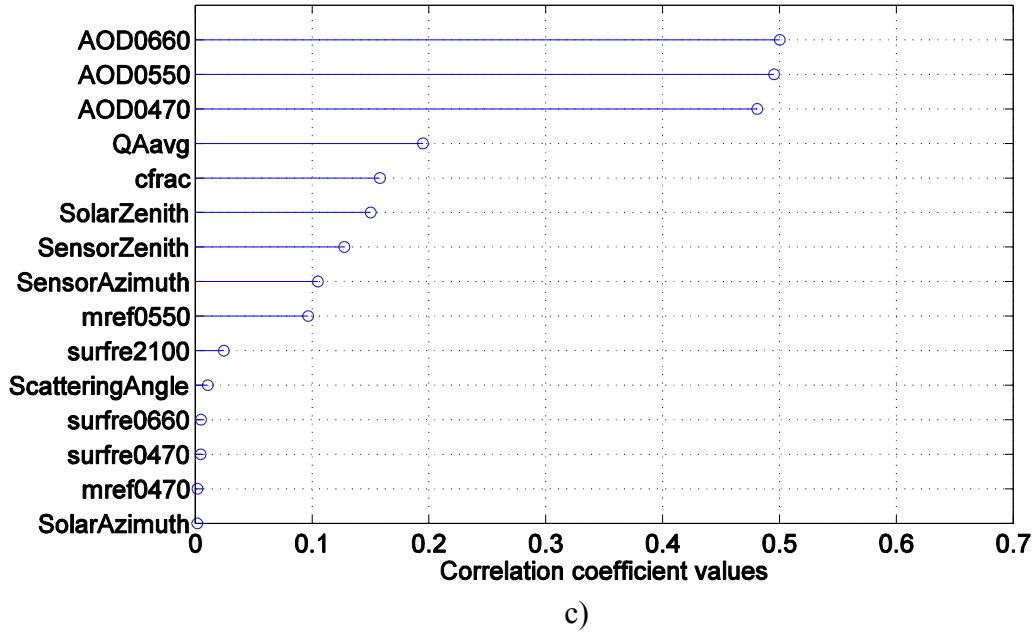

FIGURE 3. Ranking of each component affecting the bias according to the a) Bayesian MI measure $\beta = 0.05$, b) variable bin-width histogram, c) correlation coefficient

Above, we notice that the ranking of the bias-affecting components change along with the method used in the dependency analysis. It is observed that the first 2 methods are almost the same with same ranking order whereas there are significant difference between the MI estimates and the correlation coefficients, as expected. Below, we show one of the dependencies between the bias and a component, which is highly non-linear; thus dependency cannot be modeled by the linear correlation coefficient. Also, 3 clusters are noticed in the correlation ranking, whereas the ranking in different estimation techniques are closer to each other and there is no sudden jump between clusters.

## CONCLUSIONS AND FUTURE WORK

We present a brief survey on the estimation of statistical dependencies between different variables of interest. As the correlation coefficient is only optimal for linear dependencies, we focused on MI and a survey to estimate this information-theoretic quantity from data, accurately. We mentioned the difficulties in its estimation and proposed a generalized Bayesian technique. The efficient estimation of MI is applied in the detection of most important factors ending in a bias during the AOD measurements. MI estimation is compared with another approach of our team in this topic, namely Self Organizing Maps [12]. Future research will focus on the generalization of MI and TE to multivariate cases.

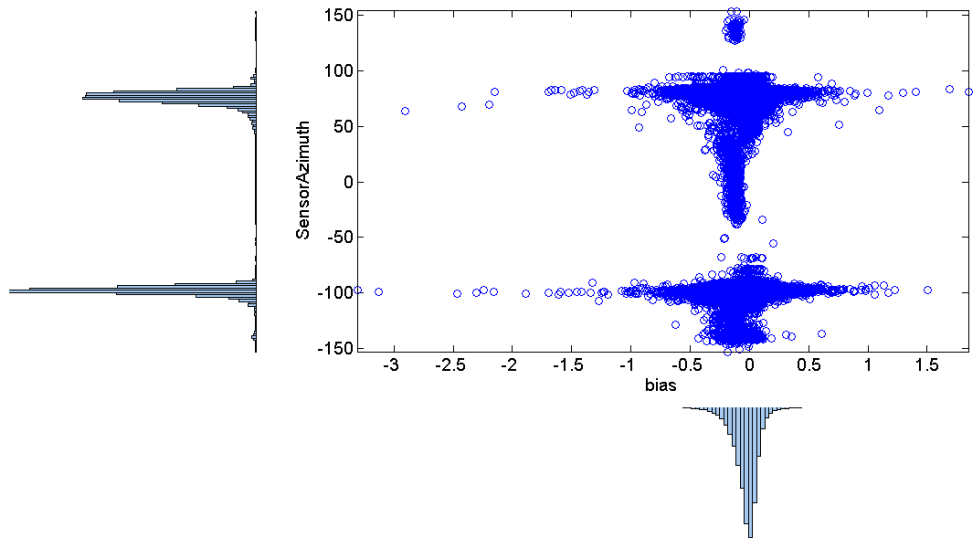

FIGURE 4. Scatter plot between the bias and the sensor azimuth (nonlinear depencency)

## ACKNOWLEDGMENTS

We thank the following agencies for funding: NASA for support under grant NNX10AM94G, DoD TATRC under grant W81XWH-11-2-0165, and the Institute for Integrative Health. We also thank Dr. Petr Tichavsky for his online code for the adaptive method.